\documentclass[conference]{IEEEtran}
\IEEEoverridecommandlockouts
\usepackage{cite}
\usepackage{amsmath,amssymb,amsfonts}
\usepackage{algorithmic}
\usepackage{graphicx}
\usepackage{textcomp}
\usepackage{xcolor}
\usepackage{comment}
\usepackage{multirow} 
\usepackage{footnote}
\usepackage{verbatim}
\usepackage{graphicx}
\usepackage[hyphens]{url}
\usepackage{listings}
\usepackage{amsmath}
\usepackage{hyperref}
\usepackage{pslatex}
\usepackage{amsfonts}
\usepackage{placeins}
\usepackage{longtable}
\usepackage{acronym}
\usepackage{subfigure}
\usepackage{stfloats}

\def\BibTeX{{\rm B\kern-.05em{\sc i\kern-.025em b}\kern-.08em
    T\kern-.1667em\lower.7ex\hbox{E}\kern-.125emX}}

\usepackage{times}
\usepackage{latexsym}

\usepackage{breqn}
\usepackage{amsmath}
\usepackage{url}

\usepackage[T1]{fontenc}

\usepackage[utf8]{inputenc}

\usepackage{microtype}
\usepackage{comment}

\usepackage{soul}
\usepackage{epstopdf}
\usepackage{booktabs}
\usepackage{graphicx}
\begin{document}

\title{Exploration of Masked and Causal Language Modelling for Text Generation \\
\thanks{ 
We are grateful for the grant “Integrating hospital outpatient letters into the healthcare data space” (EP/V047949/1; funder: UKRI/EPSRC).
\\ $^*$Corresponding Author
}
}

\author{
         Nicolo Micheletti,  Samuel Belkadi, Lifeng Han$^*$,  
           Goran Nenadic \\
         Department of Computer Science, The University of Manchester, UK \\ 

         \\ {\tt \{nicolo.micheletti, samuel.belkadi\}@student.manchester.ac.uk} 
         \\
         {\tt
         \{lifeng.han, g.nenadic\}@manchester.ac.uk}       
}

\maketitle

\begin{abstract}
Large Language Models (LLMs) have revolutionised the field of Natural Language Processing (NLP) and have achieved state-of-the-art performance in almost every task in this field. However, the prevalent approach used in text generation, Causal Language Modelling (CLM), generates text sequentially from left to right and thus limits the freedom of the model by restricting when and where each token is generated. In contrast, Masked Language Modelling (MLM), primarily used for language understanding tasks, can generate tokens anywhere in the text and in any order. This paper conducts an extensive comparison of MLM and CLM approaches for text generation tasks. To do so, we pre-train several language models of comparable sizes on three different datasets, namely (1) medical discharge summaries, (2) movie plot synopses, and (3) authorship verification datasets. 
To assess the quality of the generations, we first employ quantitative metrics and then perform a qualitative human evaluation to analyse coherence and grammatical correctness. In addition, we evaluate the usefulness of the generated texts by utilizing them in three downstream tasks: (1) Entity Recognition, (2) Text Classification, and (3) Authorship Verification. 
The results show that MLM consistently outperforms CLM in text generation across all datasets, with higher quantitative scores and better coherence in the generated texts. The study also finds \textit{no strong correlation} between the quality of the generated text and the performance of the models on the downstream tasks. Finally, we demonstrate that MLM has great potential for future research on text generation and provide direction for future studies in this area.


\end{abstract}

\section{Introduction}
\label{sec:aims_and_objectives}
The rise of Large Language Models (LLMs) has gained much attention in recent years. These models have shown remarkable ability in complex reasoning and generating human-like text. However, it became apparent since its very introduction that the transformer architecture has numerous limitations. In fact, the attention mechanism \cite{Bahdanau2014NeuralMT}, a core component of transformers, struggles with capturing long-range dependencies and does not scale efficiently to capture longer sequences. This finding led researchers to explore various ways to overcome these limitations and introduce many transformer variants \cite{49533, Dai2019TransformerXLAL, Child2019GeneratingLS}. This collective effort also led to the exponential growth in size of these models. 

Despite the tremendous effort put into language generation research, the prevalent and "almost unchallenged dogma" in text generation is that models must produce language sequentially, uni-directionally, and from left to right. The core idea is that a model would take an input text or prompt, e.g. see Table \ref{tab:clm_generation}, and then iteratively generate the next token until completion.

\begin{table}[ht]
\centering
\begin{tabular}{|l|l|}
\hline
\textbf{Step} & \textbf{Generated Text} \\
\hline
Prompt & The quick brown fox \\
\hline
1 & The quick brown fox jumps \\
\hline
2 & The quick brown fox jumps over \\
\hline
3 & The quick brown fox jumps over the \\
\hline
4 & The quick brown fox jumps over the lazy \\
\hline
5 & The quick brown fox jumps over the lazy dog \\
\hline
\end{tabular}
\caption{An example of sequential, unidirectional, left-to-right text generation using a CLM model.}
\label{tab:clm_generation}
\end{table}

This process is called Causal Language Modelling (CLM) because the probability of generating each next token is conditioned only on the tokens that appear before it in the sequence.
This unidirectional approach inevitably restricts their capacity to leverage the full context of a text, as these models are prevented from incorporating or generating tokens that appear later in the sequence. 
Howbeit, this limitation is not present when using another approach called Masked Language Modelling (MLM). 

MLMs learn to generate tokens by looking at a bidirectional context. 
Existing research has almost exclusively focused on using MLM to pre-train models for Natural Language Understanding (NLU) tasks, such as sentiment analysis, text classification, named entity recognition, etc \cite{DBLP:journals/corr/abs-1810-04805}.
However, previous research demonstrated that MLM models could be used for text generation \cite{wang2019bert} and predict tokens by considering context from both directions of a given sequence, as shown in Table \ref{fig:mlm_generation}.

\begin{table*}[ht]
\centering
\begin{tabular}{|l|l|}
\hline
\textbf{Step} & \textbf{Generated Text} \\
\hline
Prompt & The quick brown fox \texttt{[MASK]} \texttt{[MASK]} \texttt{[MASK]} \texttt{[MASK]} \texttt{[MASK]} \\
\hline
1 & The quick brown fox \texttt{[MASK]} \texttt{[MASK]} \texttt{[MASK]} \texttt{[MASK]} dog \\
\hline
2 & The quick brown fox jumps \texttt{[MASK]} \texttt{[MASK]} \texttt{[MASK]} dog \\
\hline
3 & The quick brown fox jumps \texttt{[MASK]} \texttt{[MASK]} lazy dog \\
\hline
4 & The quick brown fox jumps \texttt{[MASK]} the lazy dog \\
\hline
5 & The quick brown fox jumps over the lazy dog \\
\hline
\end{tabular}
\caption{An example of text generation using MLM.}
\label{fig:mlm_generation}
\end{table*}

Despite their potential, MLM applications in text generation, particularly for tasks such as synthetic data generation \cite{belkadi2023generating}, have been greatly overlooked, leaving their generative capabilities massively unexplored.
In this paper, we aim to challenge this assumption in \textit{text generation} by \textit{comparing CLM with MLM using models of comparable size}. By exploring this potential, we aim to find new directions for this NLP area and understand the conditions under which MLM can shine at text generation. 

Because very little research has been conducted on this topic, we decided to base our work on \textit{synthetic data generation} to allow easier evaluation by comparing the synthetic dataset with the original one, as well as to use synthetic data for the training and evaluation of our models on downstream tasks.

We raise the following research questions:
\begin{enumerate}
    \item Can MLM outperform CLM in text generation, and in what contexts?
    \item Does using domain-specific knowledge for pre-training and fine-tuning increase the quality of generated texts with regard to the quantitative metrics?
    \item Does higher quality in the texts translate to better performance on downstream tasks?
    \item What type of tokens provide enough context to generate high-quality texts?
\end{enumerate}
To answer these questions, we aimed to thoroughly evaluate and compare the ability of CLM and MLM models of comparable sizes to generate texts.
The following are the main tasks of this investigation:
\begin{enumerate}
    \item Fine-tune several Pre-trained Language Models (PLMs), including T5, BART, and BERT, and their variations.
    \item Explore different generation approaches by masking different tokens and establish which ones yield the best performance.
    \item Compare the generations of MLM and CLM models with standard quantitative metrics and qualitatively assess their generations.
    \item Examine the usefulness of fine-tuned models' generations on downstream tasks, namely Named Entity Recognition (NER), Text Classification and Authorship Verification.
    \item Interpret the results of the evaluations and conclude on the optimal types of models and masking strategies for different text generation tasks across the three datasets.
\end{enumerate}
To the best of our knowledge, this is the most extensive investigation that involves the \textbf{comparison of MLM and CLM models for text generation}.

\section{Background and Related Work}
\label{sec:background}

\subsection{Backbone Models}
In this section, we introduce the models that we used in our experiments. These models are different both in their architectures and the domains they were trained on. This will allow us to thoroughly compare how their different characteristics affect the quality of their generations.

\subsubsection{BERT family of models}
Bidirectional Encoder Representations from Transformers (BERT) \cite{DBLP:journals/corr/abs-1810-04805}. 
BERT is designed to pre-train deep bidirectional representations by conditioning on both left and right contexts in all layers. As a result, the pre-trained BERT model can be fine-tuned with just one additional output layer to create state-of-the-art models for a wide range of tasks, such as question answering \cite{Zaib2021BERTCoQACBC} and natural language inference \cite{Gajbhiye2021ExBERTAE}, without substantial task-specific architecture modifications. 
The model has 345M parameters and counts multiple variations, such as the ones introduced below.

\textbf{RoBERTa.}
Robustly Optimised BERT Pretraining Approach (RoBERTa) \cite{liu2019roberta} builds upon BERT by modifying key hyper-parameters in its pre-training process, which includes training the model longer, with more data and on bigger batches. This model showed better performance than BERT across a wide range of tasks.  The model has 355M parameters.

\textbf{BiomedNLP-PubMedBERT.}
\label{subsec:pubmedbert}
BiomedNLP-BiomedBERT-large-uncased-abstract (PubMedBERT) \cite{pubmedbert} is a BERT variant fine-tuned for the biomedical domain that uses a vast corpus of biomedical texts from PubMed, making it more effective in handling the biomedical language. The model showed state-of-the-art performance on the Biomedical Language Understanding and Reasoning Benchmark (BLURB).
We decided to use this model in order to understand whether \textit{task-specific knowledge} increases the quality of the generated texts for this specific task. \\

\subsubsection{T5 family of models}
\label{sec:t5_model}
Text-to-Text Transfer Transformer (T5) \cite{10.5555/3455716.3455856T5} is an encoder-decoder model that frames all NLP tasks as sequence-to-sequence problems. It is pre-trained on the Colossal Clean Crawled Corpus (C4) using a denoising objective, where the model learns to reconstruct corrupted input sequences. 
Providing task-specific prefixes to the input text allows it to be fine-tuned for various tasks, such as translation, summarization, and question-answering. The t5-large version is made up of 770M parameters, which is comparable to the size of large models in the BERT family. The following variation of T5 is used in our work.

\textbf{SciFive-large-Pubmed\_PMC.}
SciFive \cite{DBLP:journals/corr/abs-2106-03598} is a version of the T5 model fine-tuned on scientific and biomedical literature, specifically texts taken from PubMed and PMC (PubMed Central). SciFive is excellent in tasks related to scientific text, such as document summarisation, entity extraction, and knowledge graph construction. 
The model is used primarily for understanding and generating scientific text. Similarly to PubMedBERT, we used this model to evaluate whether \textit{task-specific knowledge} affects the generations' quality.

\subsubsection{BART}
BART (Bidirectional and Auto-Regressive Transformer) \cite{DBLP:journals/corr/abs-1910-13461} is a generalized pre-training model based on the Transformer architecture. This model counts 406M parameters and was pre-trained using five techniques: 
\begin{itemize}
\item Token masking randomly masks tokens (in the same way as for BERT).
\item Sentence permutation randomly shuffles the sentences of a document.
\item Document rotation rotates the sentence so that it begins with a randomly selected token.
\item Token deletion takes the original sentence and randomly deletes a token from the sequence.
\item Text infilling works in the same way as token masking but masks word sequences instead of single words.
\end{itemize}

\subsection{Data Augmentation}

Data augmentation has recently been used more extensively in NLP due to more work performed in low-resource domains \cite{belkadi2023generating} and the increasing popularity of large neural networks that require massive amounts of training data \cite{50311}. Applying it to this field is far more challenging than in others due to the discrete nature of language, where slight modifications can significantly alter the semantics of a sentence. Notwithstanding these challenges, numerous works have applied data augmentation for NLP. These can be simple approaches, often just involving rule-based techniques that can offer minor improvements \cite{li-etal-2017-robust}. More complex approaches are often model-based and have been applied to numerous sub-fields of NLP, such as text classification \cite{10.1145/3544558}, question answering \cite{yang2019data}, and machine translation \cite{thi-vinh_ngo_efficient_2022}.

\subsection{Synthetic Data Generation}
In contrast to data augmentation, which relies on bootstrapping existing data, synthetic data generation creates entirely new datasets from scratch. What this method essentially does is generate a fresh new dataset that has the same properties and distribution as the real-world dataset. Generative AI and Synthetic Data Generation started when Variational Autoencoders (VAEs) \cite{Kingma2013AutoEncodingVB} and Generative Adversarial Networks (GANs) \cite{goodfellow_generative_2014} were introduced, demonstrating the ability to generate realistic images, signals and tabular data.

Synthetic data in NLP became necessary as the performance of  Language Models was shown to be limited when evaluated on domain-specific benchmarks. ClinicalT5 \cite{lu-etal-2022-clinicalt5}, for example, was a version of T5 that was fine-tuned on the medical corpus. It showed better performance than T5, which was initially just trained on general text data. Most specialised domain datasets are challenging to access, showing the importance of synthetic data becoming a proxy of the original datasets for model fine-tuning \cite{belkadi2023generating}. 


\subsection{Masked Language Modeling for Text Generation}
The idea that BERT \cite{BERT_ex1} could be used for text generation was first introduced in 2019 \cite{wang2019bert}. This approach generates some (or all) the tokens simultaneously by looking at a bidirectional context. Non-autoregressive text generation was primarily used in Neural Machine Translation (NMT) to speed up inference time, by decoding multiple tokens simultaneously, at the cost of sacrificed translation accuracy compared to its counterpart, autoregressive generation \cite{10129160}. However, very few works have explored the potential of BERT in generating text outside of machine translation \cite{xiao2024are, liang-etal-2023-open}. In fact, these works on instruction following and story generations still focused on improving inference speed rather than generation quality. In addition, they did not use any domain-specific dataset and did not attempt to utilize the generations in downstream tasks. In our work, we aim to expand on this by examining the conditions under which MLM models can generate better texts than CLM models, thereby finding the best direction for future research in MLM for text generation.

\section{Methods and Experimental Design}
\label{sec:methods_and_experiments}

\subsection{Problem Formulation}
Let $C$ be a space of text features representing individual texts, such as single tokens. Let $L$ be a set of unmasked tokens, which is our context for generating synthetic data. Thereby, we have a dataset $D_{\mathcal{L}}^C$ containing a set of texts.

For each combination of tokens $\mathcal{L} \subseteq L$, we initially have a subset of data $D_{\mathcal{L}}$ defined as $D_{\mathcal{L}} = \{D_{\mathcal{L}}^n\}_{n=1}^{N_{\mathcal{L}}}$ containing texts associated with the token combination $\mathcal{L}$. Individual texts are indexed by $n$ for each $\mathcal{L}$, where $N_{\mathcal{L}}$ is the number of texts for the token combination $\mathcal{L}$.

Our objective is to generate a synthetic dataset of texts that replaces the real dataset entirely, conditioned on the combinations of relevant tokens $\mathcal{L}$. To achieve this, we aim to learn a density function $\hat{d}\{C | \mathcal{L}\}$, approximating the true distribution $d\{C | \mathcal{L}\}$ of the clinical instructions conditioned on each token combination $\mathcal{L}$.

Once the distributions are learned, we generate synthetic sub-datasets by drawing random variables from $\hat{d}\{C | \mathcal{L}\}$ for each $\mathcal{L}$. We then combine the sub-datasets to form our final dataset, which effectively replaces the original dataset while preserving its essential characteristics.

\subsection{Datasets}

\subsubsection{Medical Dataset}
For the medical task, we used the Medical Information Mart for Intensive Care (MIMIC-III) database  \cite{Johnson2016-tr, Johnson2020-ik}; in particular, we focused on the section used for the National NLP Clinical Challenges (n2c2) 2018 shared task data on adverse drug events and medication extraction with gold labels \cite{Henry2019}. We then extracted the discharge summaries for each clinical note from the dataset. We decided to use this specific dataset because it contains many caregiver notes and medication prescriptions for various clinical conditions and treatments, enabling our model to generate diverse samples. \\

\subsubsection{Movie Plot Synopsis Dataset}
\label{sec:movie_dataset}
For the second task, we used the Movie Plot Synopses with Tags (MPST) Corpus \cite{kar-etal-2018-mpst}. This dataset comprises a corpus of 14K movie plot synopses and is associated with 70 tags per synopsis, all retrieved from Wikipedia and IMDb.

\subsubsection{Authorship Verification Dataset}
The PAN Author Verification dataset\footnote{\url{https://pan.webis.de/clef23/pan23-web/author-identification.html}} contains texts from 112 individuals, each providing a written and spoken text of different Discourse Types on various topics and varying degrees of formality. These texts were transcribed into essays, emails, interviews, and speech transcriptions. All the texts in the dataset were obtained from native English speakers.

\subsection{Training and Generation}
\subsubsection{MLM Models}\text{}

\begin{figure}[ht!]
  \centering
  \includegraphics[width=.49\textwidth]{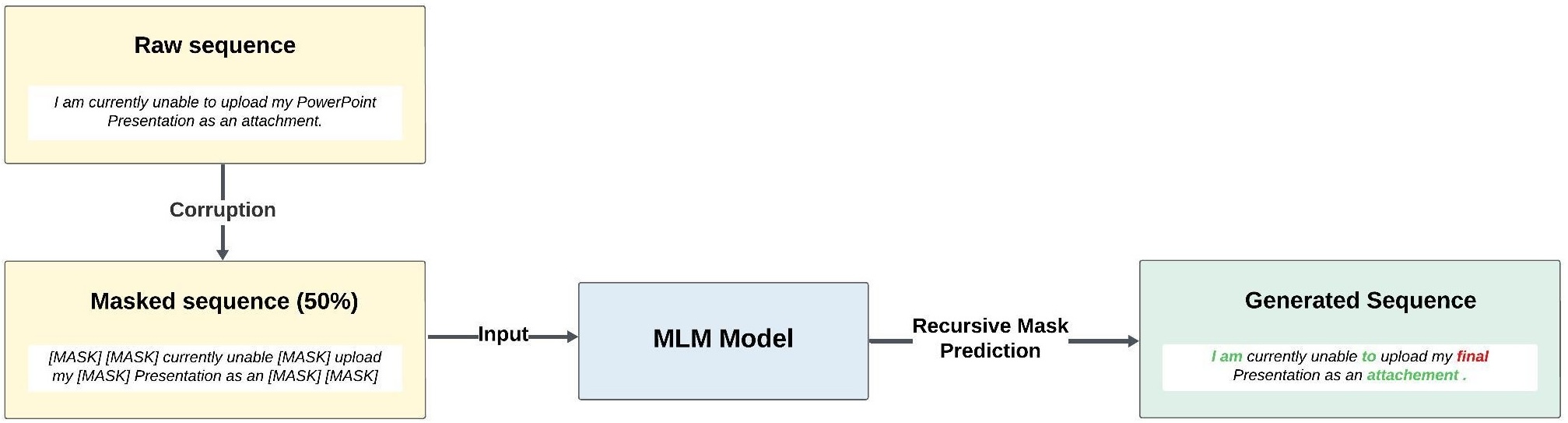}
  \caption{Examples of Inputs and Outputs for MLM models.}
  \label{fig:mlm-io-examples}
\end{figure}

\textbf{Training.} 
We used the standard MLM training protocol to train the MLM models. This approach entails masking random tokens and then letting the model learn to predict these based on the provided context, i.e., the non-masked tokens. While typically the proportion of masked tokens in the training phase is about 15\% \cite{DBLP:journals/corr/abs-1810-04805}, in our case, we set the proportion to 50\% random tokens to let the model learn in conditions that more closely align with the generation phase. An example is displayed in Figure \ref{fig:mlm-io-examples}.

\textbf{Generation.}
The process we implemented for generating texts using the MLM models consists of two steps. 
First, the original text would be corrupted according to the specific corruption method described in Section \ref{sec:generation_approaches} to contain masked and unmasked tokens.
Second, the model would predict each [MASK] token one by one until no more mask tokens remained in the text. As shown in Table \ref{fig:mlm_generation}, the model does not need to follow a particular order to unmask each token. Instead, it predicts for each [MASK] token in the order of highest confidence.

\subsubsection{CLM Models}\text{}

\begin{figure}[ht!]
  \centering
  \includegraphics[width=.48\textwidth]{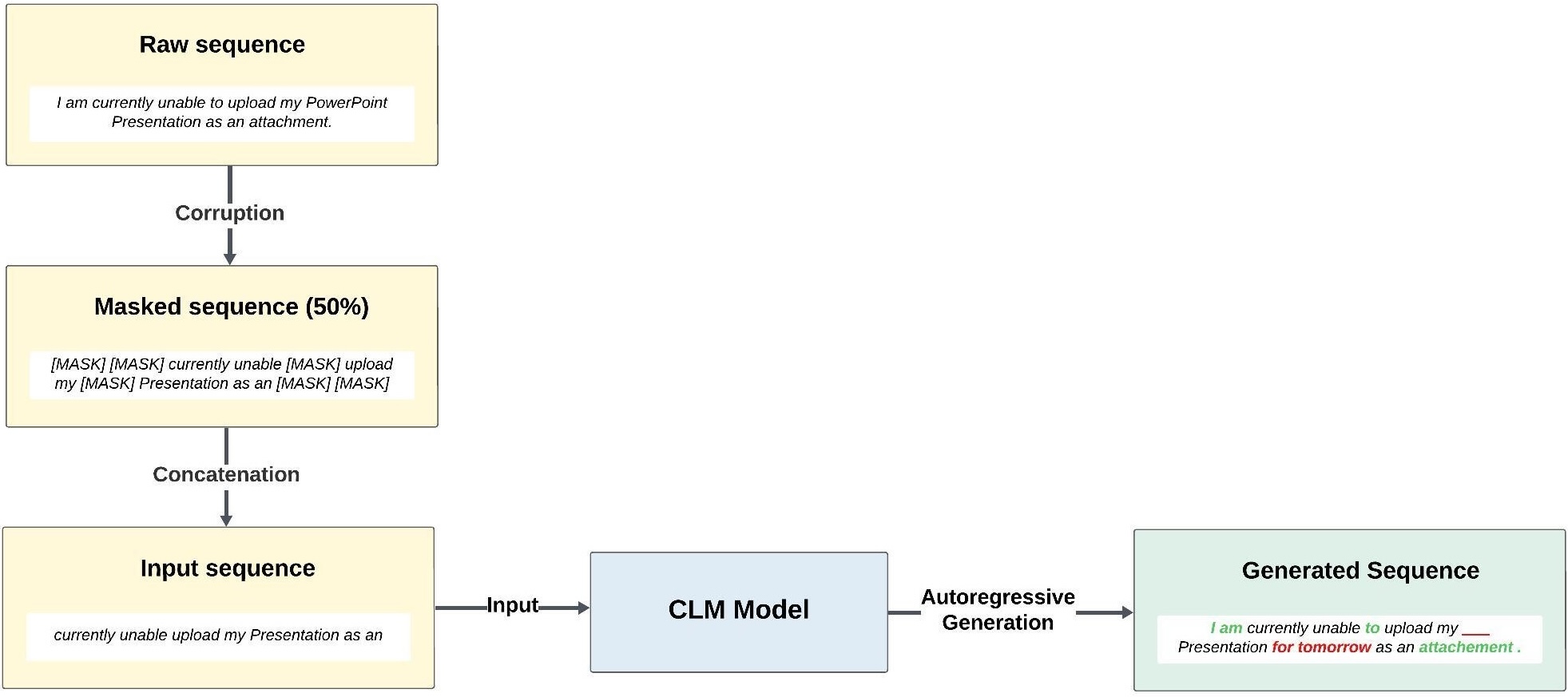}
  \caption{Examples of Inputs and Outputs for CLM models.}
  \label{fig:clm-io-examples}
\end{figure}

\textbf{Training.}
In order to train the CLM models, we followed a similar approach to that of MLMs, but removed all of the [MASK] tokens from the corrupted text and only input the unmasked tokens into the model. The latter is then trained to generate text given the list of unmasked tokens. An example can be found in Figure \ref{fig:clm-io-examples}.

\textbf{Generation.}
The generation works analogously to training: we first select the input tokens and then let the model generate the text given that context. \\

\subsubsection{Masking Techniques}
\label{sec:generation_approaches}
To select the best tokens to be the input for the generations, we employed different approaches and assessed their impact on the quality of the generations. We used the same approaches for both MLM and CLM models with the only exception being that [MASK] tokens were removed for CLM models as explained above.

The masking approaches employed are listed below:
\begin{itemize}
    \item Random: The tokens were picked randomly from the input text. With this approach, we masked a percentage of tokens from 10\% to 100\% \footnote{For CLM models, we retained that the evaluation of the 100\% masking ratio was unfair, so we excluded that from the final evaluation.}. The aim was to establish the baseline for text generation and to determine the extent to which the ratio of the number of masked tokens can impact the quality of the generated text.
    
    \item Stopwords: All tokens that are not English stopwords were masked. The aim was to understand whether punctuation alone could provide enough context to generate high-quality text.
    
    \item Punctuation: Same as above, except that we masked everything except punctuation.
 
    \item Stopwords and Punctuation: We also sought to evaluate the quality of the generations when both stopwords and punctuation were included in the input text.

    \item NER: In this approach, we used SciSpacy\footnote{\url{https://allenai.github.io/scispacy/}} to detect all entities in the text and mask the rest. This approach aimed to understand how the model can vary in its generations while maintaining the correct important entities. This generation approach was applied only to the \textbf{Medical} dataset, given that it was the only one with meaningful and easily recognisable entities.

\end{itemize}

\subsection{Downstream Tasks}
\label{sec:downstream_tasks}
In the first two tasks, we aimed to assess the quality of the generated text and its usability for training other models. We wanted to determine how quantitative metrics related to the usefulness of our texts. In the third task, we aimed to identify the approach that could produce the most similar texts to the original style according to a pre-trained model. \\

\subsubsection{Named Entity Recognition (NER) Task}
By solving the NER task, we aim to locate and classify named entities mentioned in an unstructured text into categories such as person names, organizations, locations, etc.

In this task, we evaluate the performance of a NER model \textbf{trained on the generated texts} to detect medical entities.

\begin{figure*}[t]
  \centering
  \includegraphics[width=.8\textwidth]{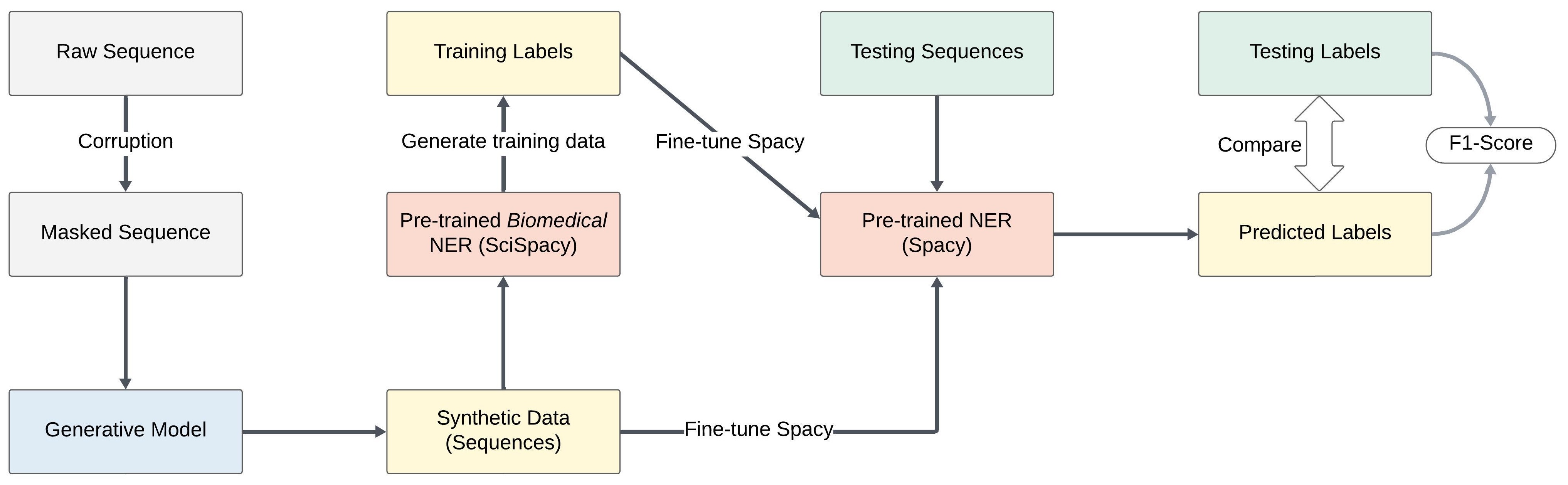}
  \caption{Experimental Design of the Downstream NER Task.}
  \label{fig:ner_design}
\end{figure*}

As shown in Figure \ref{fig:ner_design}, we first used a pre-trained NER model, named SciSpacy, to detect medical entities in the synthetic texts. The labels generated by SciSpacy would then be fed to the Spacy model along with the synthetic texts for training. Additionally, we used SciSpacy to extract entity labels from the test dataset. We then fed the testing data to the trained Spacy model and compared its predictions with the labels extracted by SciSpacy, to compute the F1 score. \\

\subsubsection{Text Classification Task}

Text classification aims to associate labels to each text of a given corpus. In our case, we perform multi-label classification, meaning that each text may have multiple correct labels.
The dataset is from the MPST Corpus, which we further processed to contain only human-generated plot synopses and reduced the set of tags to the six most frequent ones. This task focuses on the model's ability to classify texts correctly. 
As with Named Entity Recognition, we trained a BiLSTM model on the generated texts while keeping the tags unchanged. The trained models were evaluated on the same testing set to measure their movie-level F1 score, which we computed from the total number of True Positives, False Positives and False Negatives across all categories. 

\begin{figure*}[t]
  \centering
  \includegraphics[width=0.8\textwidth]{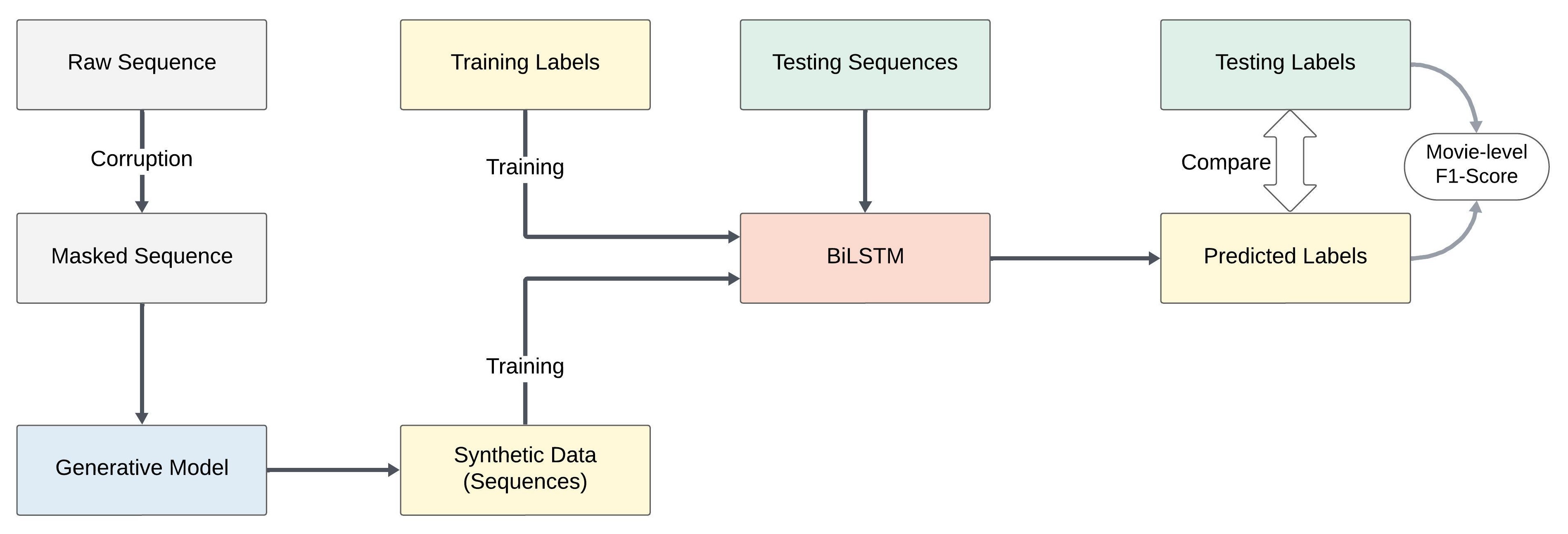}
  \caption{Experimental Design of the Downstream Text Classification Task.}
\label{fig:text_classification_design}
\end{figure*}

By following the downstream pipeline displayed in Figure \ref{fig:text_classification_design}, we aim to determine whether the texts generated by our models could be successfully used to train a text classification model. More specifically, we intent to demonstrate how much text quality matters for this task and which types of tokens provide enough context to generate useful sequences.

The BiLSTM model 
was trained on the generated plot synopses with their corresponding tags taken from the true dataset. We then used the trained model to process the test dataset and predict the most likely tags. Finally, we compared the predictions with the true testing labels and computed the movie-level F1 score. \\

\subsubsection{Authorship Verification Task}

\begin{figure*}[t]
  \centering
  \includegraphics[width=0.8\textwidth]{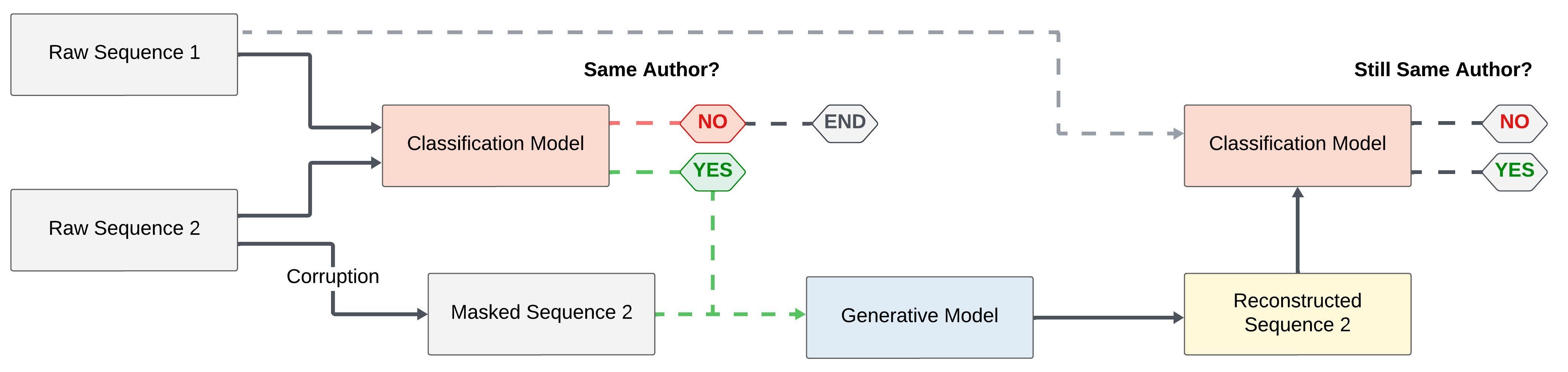}
  \caption{Experimental Design of the Downstream Author Verification Task.}
  \label{fig:author_verification_design}
\end{figure*}

Authorship verification involves the examination of linguistic patterns in two or more texts to verify whether they were authored by the same individual. Traditionally, experts conducted this analysis by considering spelling errors, grammatical inconsistencies, and other linguistic features. In recent times, however, machine learning algorithms have been trained to perform this task. As shown in Figure \ref{fig:author_verification_design}, we compiled all pairs of texts that our fine-tuned model initially identified as having the same author. Subsequently, we modified one text in each pair and assessed whether the model's prediction regarding authorship remained consistent despite these alterations.

Note that this third downstream task does not directly involve training models, but instead essentially aims to determine whether the generated text was of high quality enough to fool the predictive model into thinking that the text still comes from the same author by maintaining a similar writing style to the original text. 

\section{Results and Discussion}
\label{sec:results}
In the following section, we first show the results of the quantitative, qualitative, and downstream evaluations. We then discuss the results and their possible implications. Due to computational constraints when generating texts with MLM (as described in Section \ref{sec:limitations}), the synthetic datasets were restricted between 130 and 200 samples.

\begin{figure*}[b]
  \centering
  \includegraphics[width=0.75\textwidth]{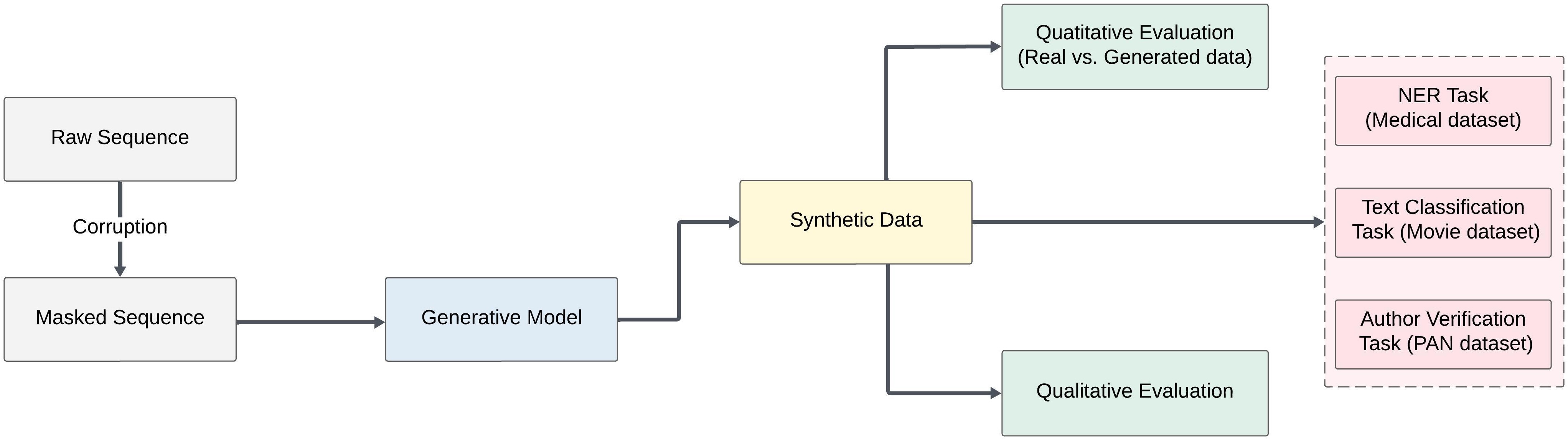}
  \caption{Evaluation Pipeline.}
  \label{fig:evaluation_pipline}
\end{figure*}

\subsection{Quantitative Evaluation}
After computing the scores for the quantitative evaluation, we visualized them in the subsequent figures. The evaluation metrics compare the generated texts derived from corrupted texts with the original texts before corruption. The figures include scores from all masking approaches discussed in Section \ref{sec:generation_approaches}. For random masking, scores are displayed as the masking ratio increases, while scores from other approaches are depicted as flat lines. The MLM models used are RoBERTa, BiomedNLP-PubMedBERT, and BERT, whereas the CLM models include T5, SciFive, and BART. \\

\subsubsection{Medical Dataset}

\begin{figure*}[ht]
    \centering
    \begin{minipage}{0.49\textwidth}
        \centering
        \includegraphics[width=\textwidth]{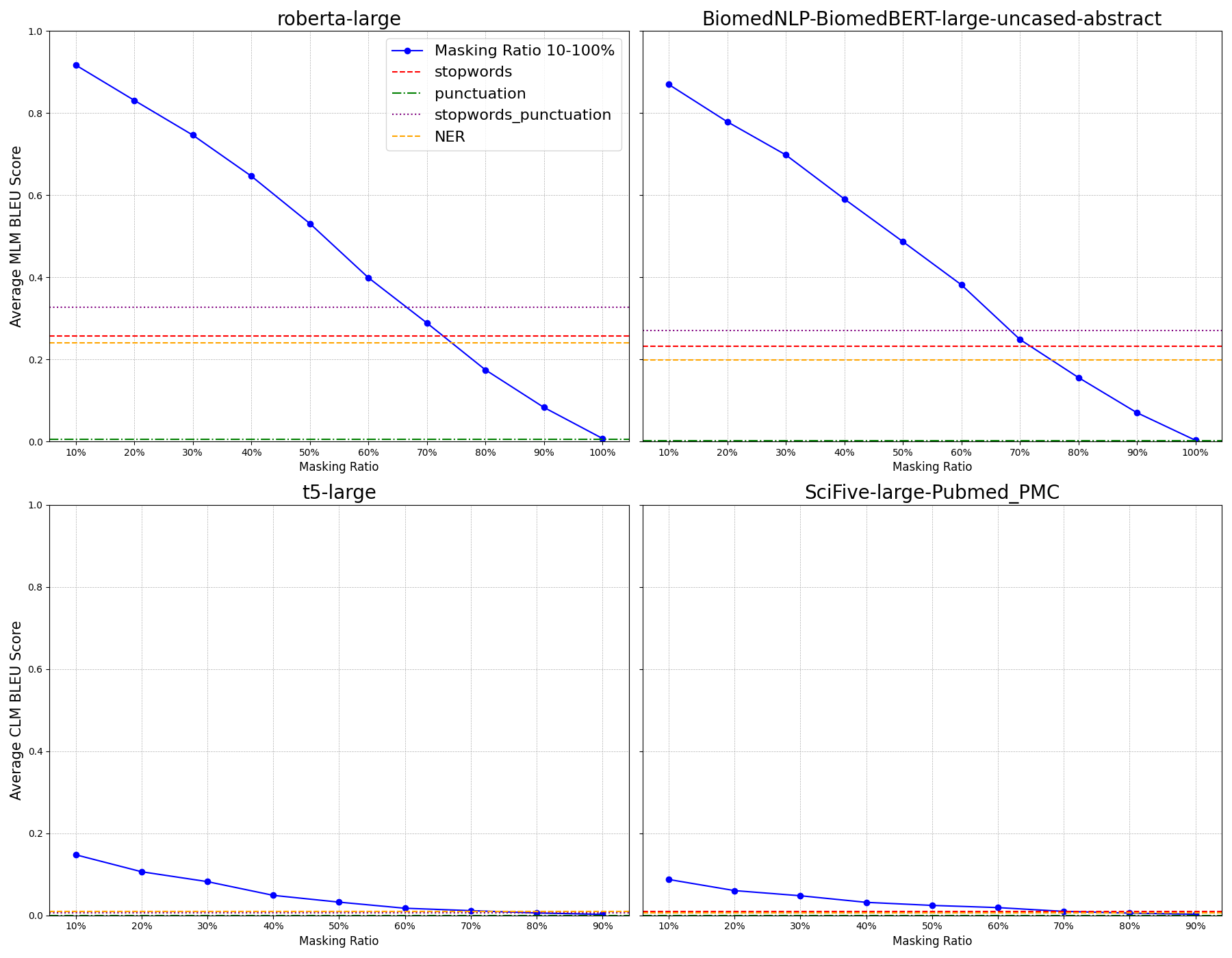}
        \caption{Medical BLEU Scores.}
        \label{fig:medical_bleu_scores}
    \end{minipage}
    \hfill
    \begin{minipage}{0.49\textwidth}
        \centering
        \includegraphics[width=\textwidth]{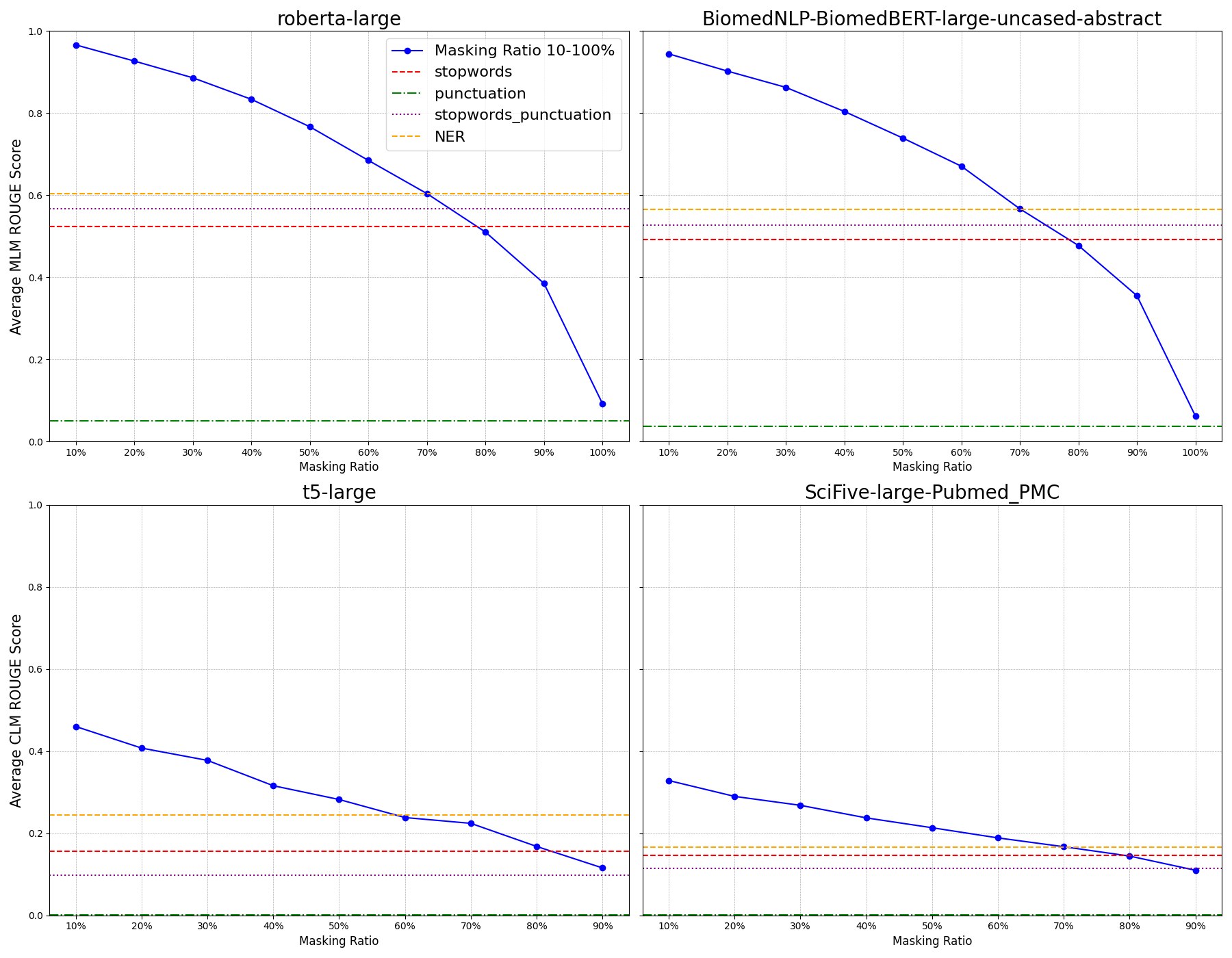}
        \caption{Medical ROUGE Scores.}
        \label{fig:medical_rouge_scores}
    \end{minipage}
    \vskip\baselineskip
    \begin{minipage}{0.49\textwidth}
        \centering
        \includegraphics[width=\textwidth]{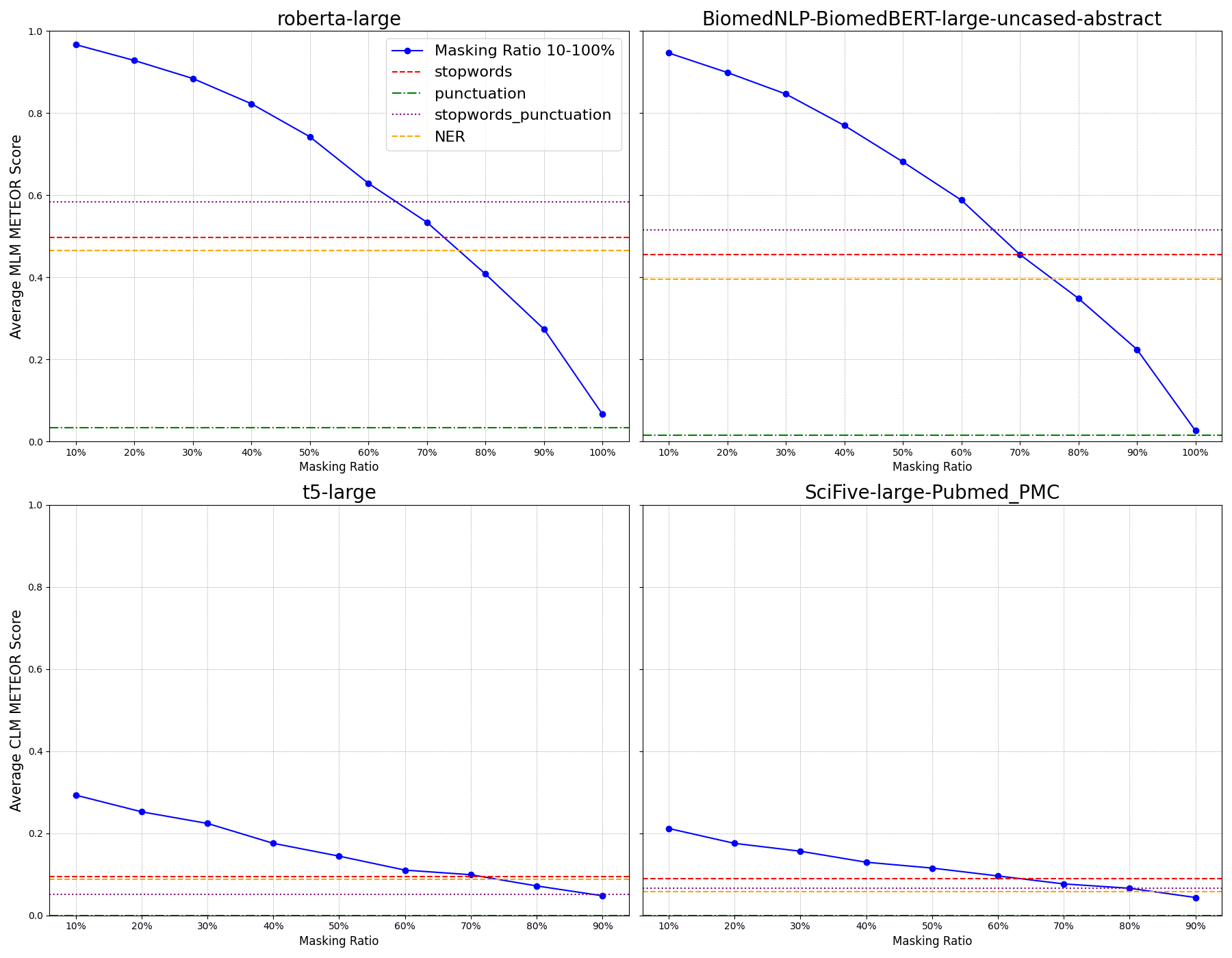}
        \caption{Medical METEOR Scores.}
        \label{fig:medical_meteor_scores}
    \end{minipage}
    \hfill
    \begin{minipage}{0.49\textwidth}
        \centering
        \includegraphics[width=\textwidth]{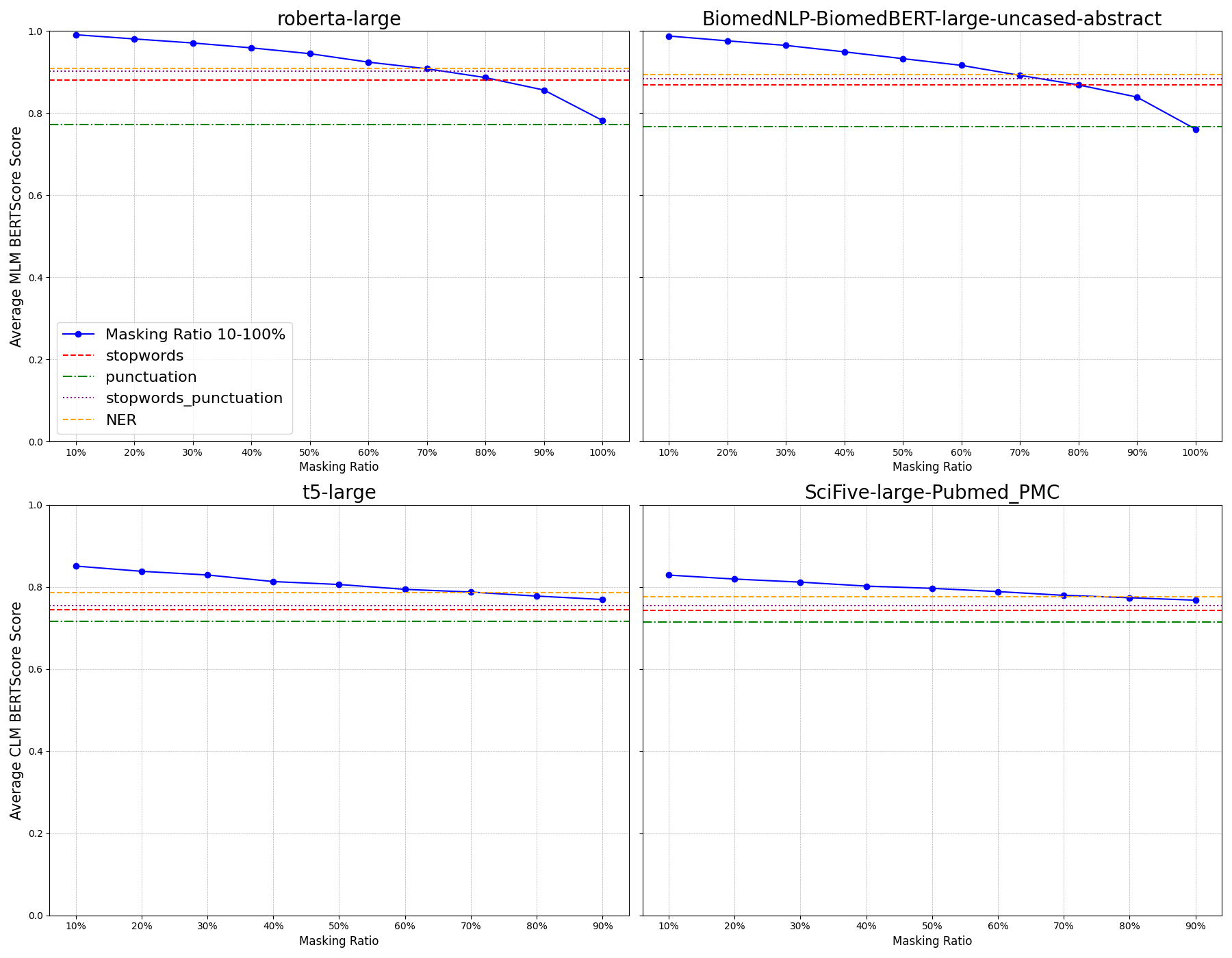}
        \caption{Medical BERT Scores.}
        \label{fig:medical_bert_scores}
    \end{minipage}
\end{figure*}

As illustrated in Figure \ref{fig:medical_bleu_scores}, the generations from the MLM models (RoBERTa-large and BiomedNLP-PubMedBERT) align more closely with the original text, thereby achieving higher scores. These models are more \textbf{sensitive} to \textit{changes in the masking ratio}, performing better when the ratio is lower, and exhibiting very low scores when the ratio reaches 100\%. In contrast, the CLM models display low scores across all generation approaches, indicating low precision as reflected by the low BLEU scores. For the MLM models, masking punctuation and stopwords results in mid-range scores, whereas for the CLM models, these approaches yield scores at the bottom of the graph.

Figure \ref{fig:medical_rouge_scores} indicates that, although still inferior to the MLM models, the ROUGE scores of the CLM models are relatively better than their BLEU scores. This suggests that at least some of the words generated by the CLM models also appear in the reference texts, indicating a higher recall.

Figure \ref{fig:medical_meteor_scores} supports this observation through the METEOR evaluation, a recall-oriented metric, which shows relatively good scores for the CLM models compared to their BLEU scores. However, CLM models generate many tokens that are not present in the reference texts, unlike MLM models.

In Figure \ref{fig:medical_bert_scores}, the BERTScore reveals that despite the specific medical language, all models can generate samples semantically similar to the reference texts. This demonstrates good adaptability to the task-specific domain for all models, with MLM models exhibiting slightly better performance. \\

\subsubsection{Movie Dataset}
As observed in Figure \ref{fig:movie_bleu_scores}, less masking results in higher scores. When generating samples from the movie dataset, the MLM model continues to achieve higher scores. Although T5-large's scores remain as low as those with the medical dataset, BART-large shows significant improvement, achieving good overall scores. Additionally, the model appears more sensitive to the level of context, suggesting it can better utilize context to produce informed generations. Neither the punctuation nor stopwords masking approaches provide sufficient context for generating high-quality samples, as evidenced by their scores being at the bottom of the graph. \\

\subsubsection{Author Dataset}
\begin{figure*}[ht]
    \centering
    \begin{minipage}{0.49\textwidth}
        \centering
        \includegraphics[width=\textwidth]{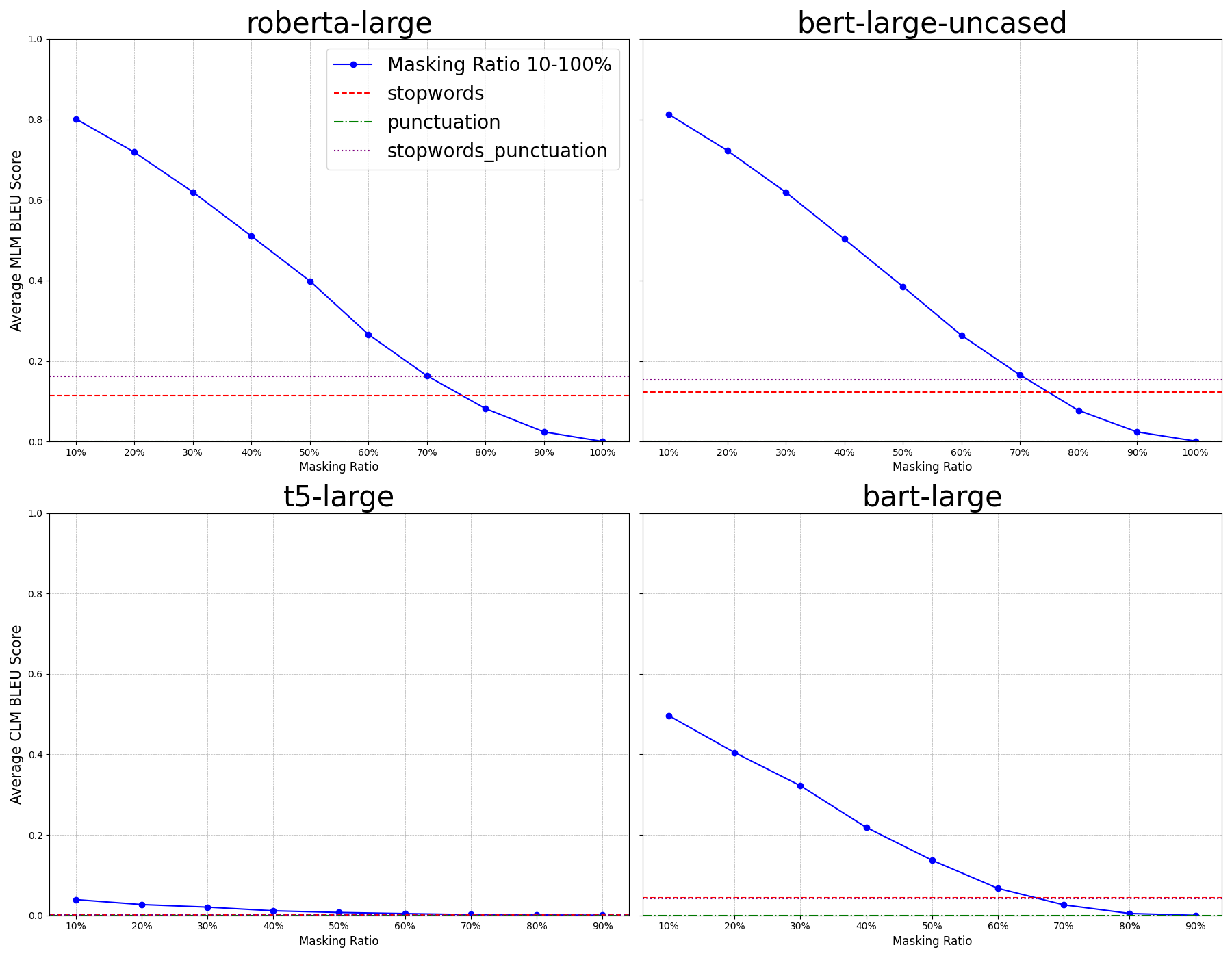}
        \caption{Movie BLEU Scores.}
        \label{fig:movie_bleu_scores}
    \end{minipage}
    \hfill
    \begin{minipage}{0.49\textwidth}
        \centering
        \includegraphics[width=\textwidth]{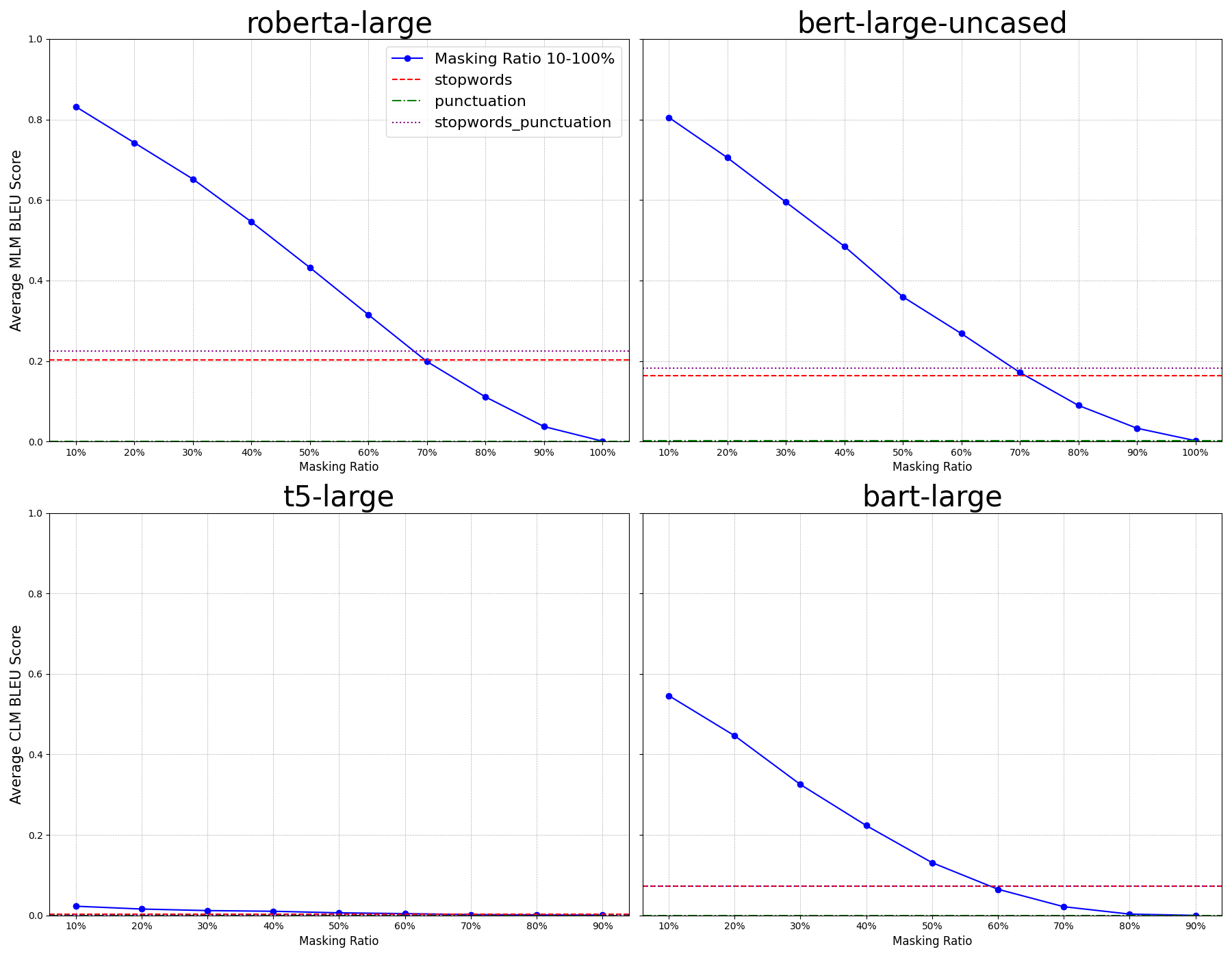}
        \caption{Author BLEU Scores.}
        \label{fig:author_bleu_scores}
    \end{minipage}
\end{figure*}

Once again, Figure \ref{fig:author_bleu_scores} demonstrates that MLM models outperform CLM models. Bart-large performs well through the variable ratios compared to the previous two datasets, while t5-large cannot produce accurate generations.

\subsection{Human Evaluation on Sampled Data}

In this subsection, we pick the first sample generated by each model using a random 50\% masking ratio. \\

\subsubsection{Medical Generations}\text{}

\textbf{MLM Generations.}
As shown in Table \ref{tab:medical_mlm_generations}, MLM sequences generated by Roberta-large are indeed of high quality. Although they show a few hallucinations, we observe a high level of coherence in the texts and sequences that match the expected length for discharge summaries. In addition, these texts show an appropriate structure for medical discharge summaries as they include relevant sections such as patient instructions, medication changes, and follow-up recommendations. What is also remarkable is that the model demonstrates a good grasp of medical terms, such as \textit{"angioedema," "pneumonia,"} and \textit{"endoscopy"}. The texts are generally coherent; however, information is sometimes introduced out of nowhere. For example, in the fourth summary, \textit{"mm nodular area on the chest CT"} is disconnected from the rest of the summary.

\begin{table}[hb!]
    \centering
    \begin{tabular}{|p{0.9\linewidth}|}
    \hline \vspace{.03cm}
        please shower daily including washing incisions gently with mild soap, no baths or swimming pat and dry at all times please no powder, powder, cream, or ointments to cover each incision please weigh yourself and measure yourself every evening and every morning, these should be written down on the chart no driving for one month or while taking narcotics. driving will be discussed at follow up appointment with surgeon please no lifting more than pounds for weeks please call with any questions or concerns telephone or fax please call cardiac surgery office with any questions or concerns telephone or fax. \vspace{.15cm} \\
    \hline
    \end{tabular}
    \caption{Roberta Medical Generations based on 50\% random context.}
    \label{tab:medical_mlm_generations}
\end{table}

\textbf{CLM Generations.}
\begin{table}[bht!]
    \centering
    \begin{tabular}{|p{0.9\linewidth}|}
        \hline \vspace{.03cm}
        be at follow up please call with any or concerns/fax cardiac surgery office any or concerns/fax cardiac surgery office any or concerns telephone. answering service contact on hours telephone. answering service contact on hours telephone. answering service contact on hours telephone. answering service contact on hours telephone. answering service contact on hours telephone. answering service contact on hours telephone. answering service contact on hours telephone. answering service contact on hours telephone. answering service contact on hours telephone. answering service contact on hours telephone. answering service contact on hours telephone. answering service contact on hours telephone.', 1: 'take any new. Please take any new of regular medications take any new as ordered. Please of regular medications take any new. Please of regular medications take any new. Please of regular medications take any new. Please of regular medications take any new of regular medications take any new... \vspace{.15cm}\\
        \hline
    \end{tabular}
    \caption{T5 Medical Generations based on 50\% random context.}
    \label{tab:medical_clm_generations}
\end{table}
In Table \ref{tab:medical_clm_generations}, the texts generated by T5-large show many \textit{repetitions} where multiple phrases are repeated over the entire length of the sequence. Note, for example, the phrase \textit{"answering service contact on hours telephone"} that is repeated twelve times over the entire sequence. 
First of all, the generated texts lack the structure and organization that proper medical discharge summaries would require. Second, most of the information appears randomly generated and without a clear flow. This all together makes it very difficult to understand the key instructions given by the summary. In addition, some texts contain incomplete sentences, such as \textit{"take any new. Please take any new or regular medications as ordered. Please, of regular medications, take any new."}. Over longer sentences, it is clear that the CLM model cannot maintain any coherence in its generated texts; indeed, while the texts include medical terms and phrases, they are often used in a disjointed way and lack context. For instance, \textit{"blood systolic. blood systolic. blood primary care doctor to hospital with a reaction called angioedema."} is a clear example of hallucination. Finally, we can also observe many instances of irrelevant information being generated, such as \textit{"if it becomes black or blood in stool, go to the hospital."}, which appears in the summary without appropriate context.

\subsubsection{Movie Generations}\text{}

\textbf{MLM Generations.}
Table \ref{tab:mlm_movie_generations} shows that the texts generated by Roberta-large have different levels of coherence and grammatical correctness. Overall, the summaries convey the main storylines smoothly and introduce the central characters at the right time; however, the narration is disrupted multiple times because of the hallucinations introduced by the model.

Many of the transitions in the summaries happened all of a sudden, which makes it harder to follow the plots. The second summary, for example, jumps from the childhood of the protagonist to his adult life. In addition, the generated text sometimes includes irrelevant details that draw attention away from the main storyline, such as the \textit{"Glock"} and the slow movement of the desk in the fifth summary, which is totally out of context. The model is sometimes limited in its ability to maintain long-term dependencies between characters in the plots; for example, in the fourth plot, the summary contradicts itself multiple times about the actions of \textit{Duke} and \textit{Meta}.

Nonetheless, the generations are mostly understandable. The main characters, settings, and conflicts are all present in the summaries. They are also easy to read, have clear language, and use languages similar to the ones used in the original training dataset. Finally, the summaries are very different in style and tone, showing that the models can adapt to different genres.

\begin{table}[hbt!]
    \centering
    \begin{tabular}{|p{0.9\linewidth}|}
    \hline \vspace{.03cm}
    The film opens with a montage of scenes of daily life in Australia during the apocalypse: a rural school in the desert, a busy street in a small town, a traffic jam in the city, all being affected by the adverse weather conditions that suddenly appear. Only the local Aboriginal people seem to recognize the changes. The film's first scene is one of these "freak occurrences" in Sydney, where it occurs among a group of men in a pub, all affected by the mysterious effects of the weather. Arriving at the court of public safety, the case is ruled not guilty; and the men are given the opportunity to use the public legal aid system, where Burton is the only legal counsel procured for their defense. The circumstances under which he was contacted and retained are unusual, in that his law practice is corporate, and not criminal defense. The film focuses on the case, and the professional and personal lives of the unravel involved, including bizarre dreams that Burton relates to the court. Interacting interviewing one of the accused (David Gulpilil), he is also able to uncover a strange ritual that has taken place in his home, and connect his rituals with the murder case, which turns out to have been an Aboriginal ritual performed by curse same shaman that the victim believed. Learning more about Aboriginal practices and the idea of Dreamtime as a parallel world of existence, Burton concludes that this strange ritual bodes ill for the apocalypse. The film ends with a confrontation between Burton and the victim's shaman at a subterranean burial site. \vspace{.15cm}\\
    \hline
    \end{tabular}
    \caption{Roberta Movie Generations based on 50\% random context.}
    \label{tab:mlm_movie_generations}
\end{table}

\textbf{CLM Generations.}
\begin{table}[h!]
    \centering
    \begin{tabular}{|p{0.9\linewidth}|}
    \hline \vspace{.03cm}
    With montage scenes of daily Australia 1970s: desert main of a jam in the, being unusually adverse weather that appear only the aboriginal to recognize the significance of weather. The Australian legal aid, (Chamberlain) defence the circumstances by which, in that taxation the and personal lives unravel.  He is plagued by bizarre dreams begins to connection to one of the accused (David Gulpilil) also feels connected increasingly strange phenomena city. His intensify along with the case, which he comes an aboriginal killing curse in which the victim about aboriginal and the as a parallel world of existence, Burton comes to weather bodes the end of the world. Overcoming the climaxes confrontation the lawyer and tribe' subterranean sacred site. After the sewer, he on beach and stares. Screen filled by the shot of a ocean wave unclear if he is witnessing or in' final, he stares at the ocean.  The film ends with one of these "freak" Sydney an occurs among a of in the mysterious of the mysterious death of one of at coroner, death ruled homicide; men.  Burton escapes to in the tunnel various relics. \vspace{.15cm}\\
    \hline
    \end{tabular}
    \caption{Bart Movie Generations based on 50\% random context.}
    \label{tab:movie_clm_generations}
\end{table}
Table \ref{tab:movie_clm_generations} shows that overall, the quality of the writings produced by Bart-large is far below that of the MLM generations. The primary issue is the lack of coherence in specific passages as the text occasionally struggles to maintain a clear narrative flow, with abrupt transitions and unclear connections between each sequence. Most passages contain vague statements that leave room for interpretation and seem more like a collection of random sentences lacking logical connections between them. Another critical issue is that some sequences contain grammatical errors, such as incorrect verb tenses, subject-verb agreement, and awkward sentence structures. Additionally, the texts sometimes lack clarity and precision in their language, with some occasional appearance of random characters. \\

\subsubsection{Author Generations}
Below are the generations of Roberta-large and Bart-large. It should be noted that some of the problems the models show with these generations were also present in the original dataset, so the training data has probably influenced the quality of the generated texts. 

\textbf{MLM Generations.}
\begin{table}[]
    \centering
    \begin{tabular}{|p{0.9\linewidth}|}
    \hline \vspace{.03cm}
    'hi, thank you so much for clarifying this, im so sorry for the inconvenience i am happy to take all the shifts for tomorrow, again thank you for your help Hi, good afternoon, i have been unable to upload my powerpoint presentation as an attachment as the file is not working i have attempted to upload it as an onedrive, however i am experiencing issues could i please upload the slides from my memory stick if any further attempts fail i will continue to continue to attempt to upload if you could let me know if this is working i would be very grateful kindest regards, Hi, i have estimated as an end date for tomorrow hope this is okay truly, Hi, i have attached pictures of my presentation, as have many others, good afternoon, i hope your well im sorry for the delayed reply, i was unable to hear back from you guys about my dbs, which has now been received i have scanned my dbs but they are much bigger than a4x5 printer normally scans so they will not show all the information on each side i was able to print out all the documents on hard drive i have attached the pdf containing the information required per the interview, there are also 2 pdfs containing my CV and email address thank you for your help, Hi, my lovelies, i hope youre all well i am so happy to inform you i have secured the role with i would not have got through the interview without all your support and help, i cant thank you enough for your help with my placement i hope you guys will be able to answer if i am missing anything in any of my questions, i can tell you that i will be completing my placement for 100 days, rather than a year anything which is very useful to carry around day i am currently searching for a diary what is training and is it something i should bringwear me as i havent been provided with a location yet,' \vspace{.15cm}\\
    \hline
    \end{tabular}
    \caption{Roberta Author Generations based on 50\% random context.}
    \label{tab:mlm_author_generations}
\end{table}
Table \ref{tab:mlm_author_generations} shows that one significant issue within texts generated by Roberta-large is the lack of proper grammar, punctuation and sentence structure. The model struggles to maintain coherence and clarity, often producing fragmented and poorly organized ideas. In addition, the texts are mostly made up of repeated phrases, lacking smooth transitions between ideas. However, the model does have the ability to capture different writing styles, such as casual conversations or more formal tones. Overall, the generated texts are understandable and do not include significant errors. 

\begin{table}[]
    \centering
    \begin{tabular}{|p{0.9\linewidth}|}
    \hline \vspace{.03cm}
    Hi thank you so clarifying that im will bring the for once thank you, good afternoon, i am unable to my powerpoint presentation as attachment as the is too big to it as onedrive i am experiencing some issues with my memory, could i the presentation memory all other attempts fail i will however to you this is i be very grateful kindest regards, date placement i hope yours, hi have attached dbs thanks,    Hi ,  I have attached my dbs  Thank you for your support,  Hi,  I hope youre well, I have a queries the placement hope will be answer if thats ok to of my questions can tell you that i be for days over a will to carry day to day i am currently what training like should i been with location yet so routes would be helpful recommendations the prices on number days i will completing you guys before placement any you think relevant for to know i contacted this again thank you for my support, thank you enough   Good afternoon,  My name is   and I am a student at  , I am on placement for   with   I am so pleased to the role with i not gone through the application process as i have not received any support, Thank you enough,  Hello,  your have attached the dbs to my presentation   Thanks,  Good morning,  my university about form has now i have as bigger than a4 only a4 show all each am to documents as copies 4 the required per there are also 2 attached containing address \vspace{.15cm}\\
    \hline
    \end{tabular}
    \caption{Bart Author Generations based on 50\% random context.}
    \label{tab:clm_author_generations}
\end{table}

\textbf{CLM Generations.}
Table \ref{tab:clm_author_generations} displays texts generated by Bart-large. As we can observe, the sequences cover various topics and styles, including personal communication, academic writing, and conversational transcripts. However, the overall quality of the generated content is poor and lacks coherence, proper grammar, and punctuation. As with the movie synopses, the model struggles to maintain clarity, generating fragmented sentences. The information presented is often confusing and contains random or irrelevant details. Overall, the models fail to maintain a core idea, demonstrating a lack of coherent flow throughout their generations.

\subsection{Downstream Evaluation Tasks}
In this section, we evaluate the generated texts by utilizing them in three downstream tasks, namely NER, Text Classification, and Authorship Verification. These results will suggest the relationship between \textit{the quality of the generations and their usefulness} in a practical setting, assessing the potential to use them as synthetic datasets for different NLP tasks. The design of each downstream task was explained in Section \ref{sec:downstream_tasks}. \\

\subsubsection{Downstream NER Task}

\begin{figure*}[ht]
  \centering
  \includegraphics[width=0.6\textwidth]{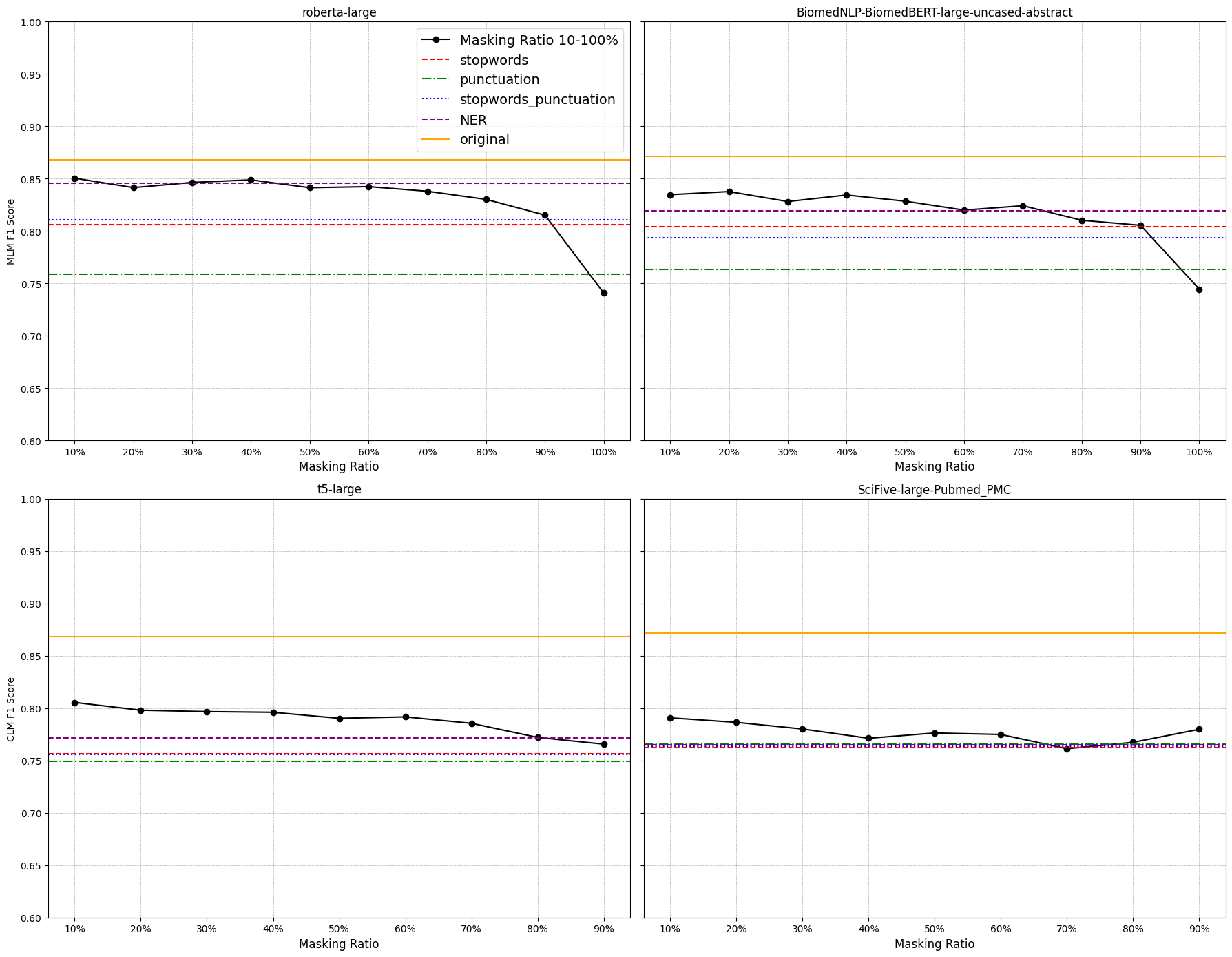}
  \caption{Downstream NER Evaluation Results.}
  \label{fig:downstream_ner_results}
\end{figure*}

Figure \ref{fig:downstream_ner_results} demonstrates that, although models trained on real data perform slightly better (orange line), models trained on generated data could perform decently. The graph exhibits a similar trend to the quantitative metric graphs, showing high scores for higher ratios and a less decisive decline with reduced contexts. Unlike the quantitative metrics, synthetic data generated using stopwords and punctuation as context could train the models to achieve performance comparable to random masking. For MLM models, NER tokens could provide sufficient context to train the model effectively.

Remarkably, even when the masking ratio reaches 100\%, the F1 score, though sharply decreased, remains effective enough to train the model, achieving an F1 score of nearly 0.75. This contrasts with the low quantitative scores observed at this ratio. 
Therefore, these results suggest that for training NER models, the quality of generated samples has a limited impact on performance. \\

\begin{figure*}[ht]
    \centering
    \begin{minipage}{0.49\textwidth}
        \centering
        \includegraphics[width=\textwidth]{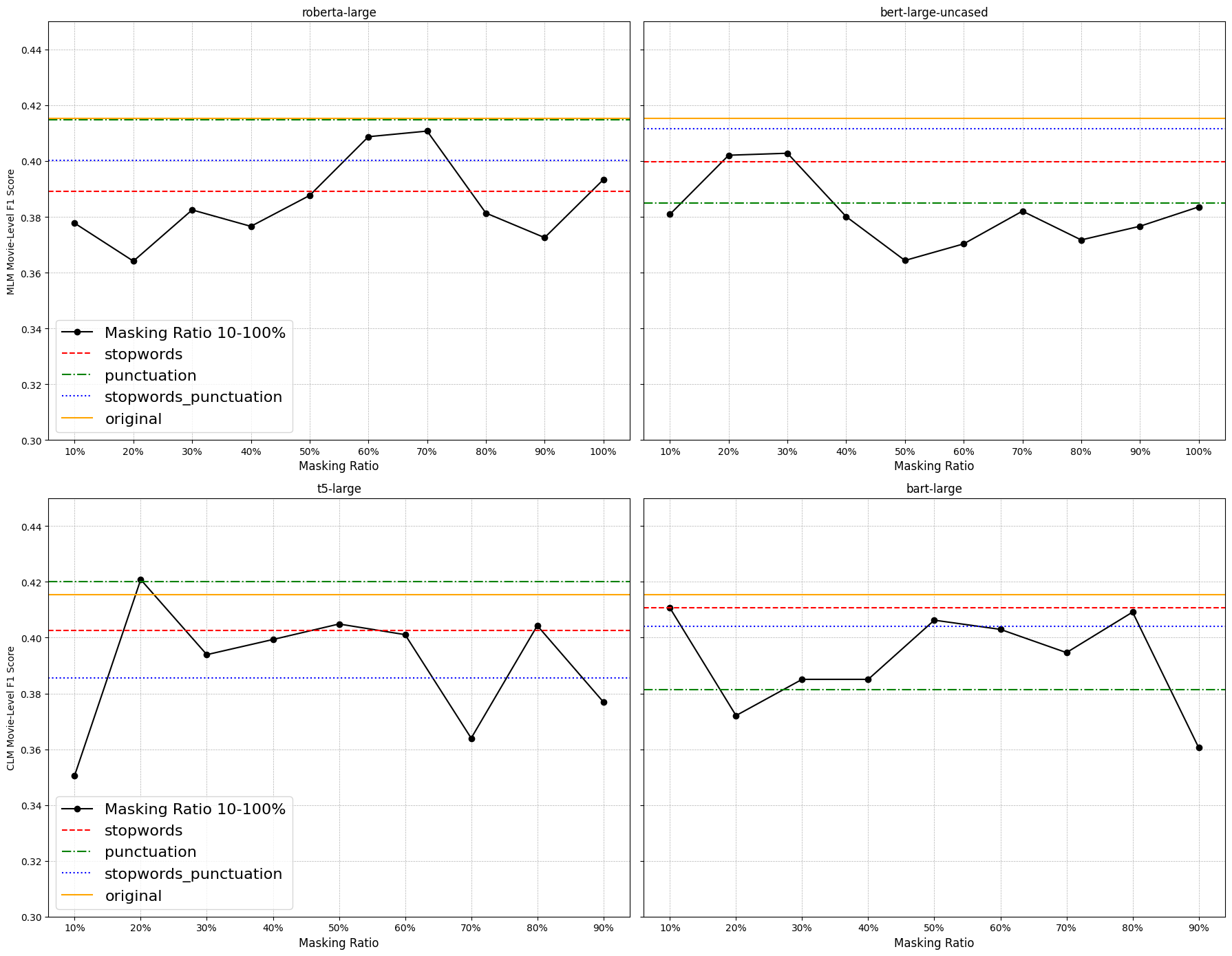}
        \caption{Downstream Text Classification Evaluation Results.}
        \label{fig:downstream_tc_results}
    \end{minipage}
    \hfill
    \begin{minipage}{0.49\textwidth}
        \centering
        \includegraphics[width=\textwidth]{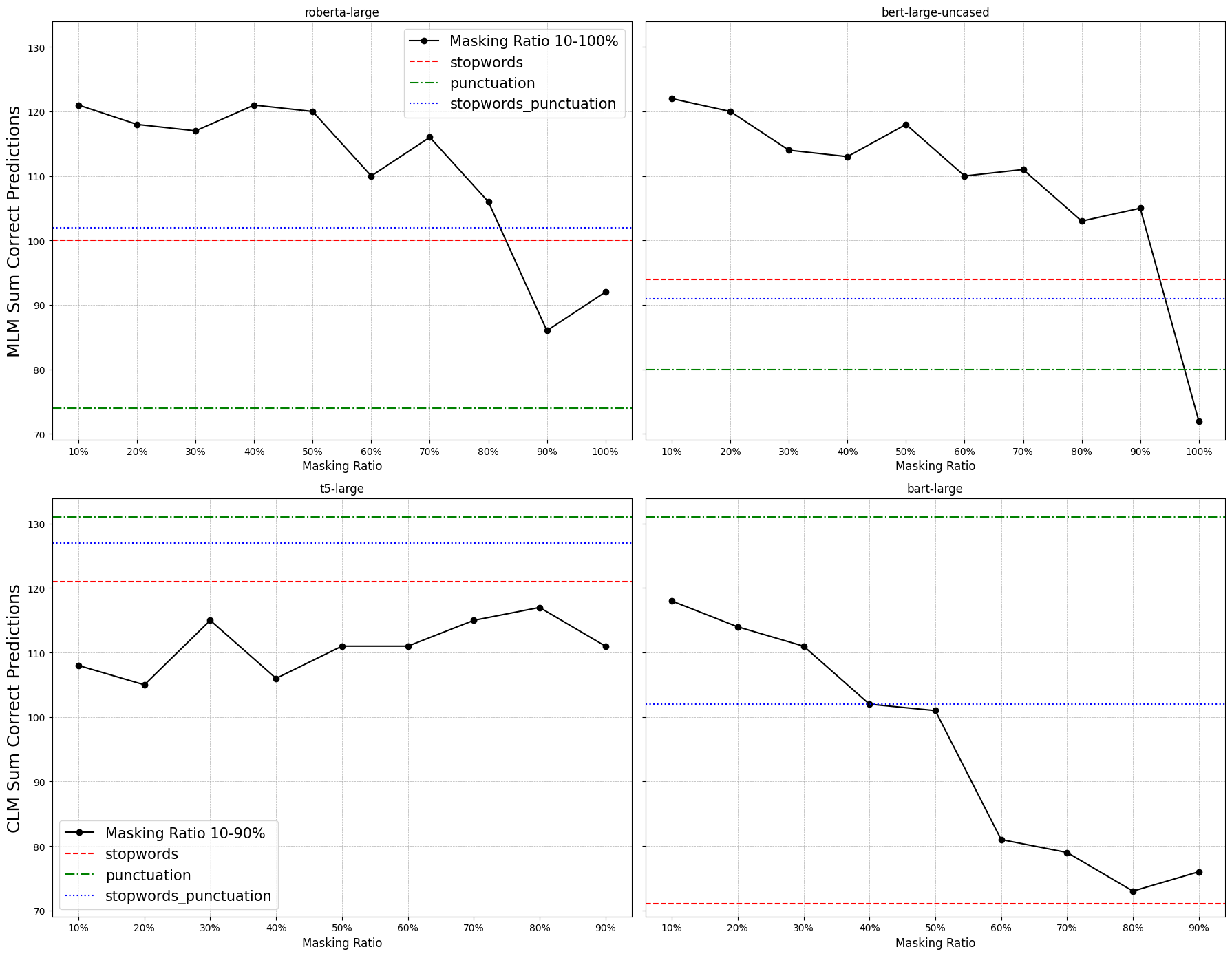}
        \caption{Downstream Authorship Verification Evaluation Results.}
        \label{fig:downstream_av_results}
    \end{minipage}
\end{figure*}

\subsubsection{Downstream Text Classification Task}
The results in Figure \ref{fig:downstream_tc_results} indicate that the models trained on generated data exhibit significant performance variation based on the quality of the generated texts. The best model achieved an F1 score of 0.42, while the worst scored 0.36. Unlike in the NER task, the quality of the text does not strongly correlate with model performance. In fact, the RoBERTa-large model with a 100\% masking ratio outperforms most other ratios, and the T5-large model with punctuation masking performs even better than when trained on the original data.

It is important to note that, due to hardware constraints, we could only train the models on 200 samples, which limits their overall performance in this task. \\

\subsubsection{Downstream Author Verification Task}
As we can observe from Figure \ref{fig:downstream_av_results}, the generated texts exhibit performances comparable to the quantitative metrics for three out of the four models. This indicates that the models can leverage the context appropriately. However, for the MLM models, performance decreases sharply at the higher masking ratios (90-100\%). In contrast, T5-large shows no correlation between the number of matches and the amount of context provided.

Interestingly, punctuation alone achieves the highest performance for the CLM models. However, upon examining the actual generations, we noted that the models exhibit clear hallucinations when given just punctuation. Specifically, all generations are merely ",,,,,", leading the prediction model to correctly predict "Yes" with low confidence.

Regarding stopwords and the combination of stopwords and punctuation, they demonstrate fairly good performance, positioning them in the middle of the graph. Punctuation alone does not provide sufficient context for the MLM models, resulting in very low performance, as depicted at the bottom of the figure.

\subsection{Discussion}
The quantitative and qualitative evaluations of this study, as well as the performance observed in the downstream tasks, provide valuable insights into the effectiveness of different language models and suggest optimal masking approaches for text generation.
Our main findings are given below: \\

\textbf{(1) MLM Outperforms CLM:} MLM consistently outperformed CLM in text generation across all three datasets. Quantitative metrics showed that texts generated by MLM models were more aligned with the original reference texts. In addition, qualitative human evaluation indicated that MLM generations had better coherence and grammatical accuracy.

\textbf{(2) Domain-Specific Knowledge does not Improve Performance:} Incorporating domain-specific knowledge in the generation models did not enhance performance. This is evidenced by the scores from BiomedNLP-PubMedBERT and SciFive-Large-Pubmed\_PMC, two models focused on the biomedical domain, which did not achieve better scores than the general models at any quantitative metrics.


\textbf{(3) There is no strong correlation between the quality of generated texts and performance in downstream tasks.} While most models trained on higher-quality generations performed well, some models trained on lower-quality data (e.g., 100\% masking ratio) also achieved decent results. This finding suggests that downstream tasks are more resilient to noise introduced by generative models. It is also possible that synthetic generations enhance model robustness by introducing significant variability in the training sets.

\textbf{(4) Effectiveness of Random Masking with Lower Ratios:} As expected, random masking with lower masking ratios (preserving more context) resulted in higher-quality generations. However, even with limited context, such as NER or stopwords\_punctuation masking, MLM models could still generate texts of good overall quality.

\section{Conclusions and Future Work}
\label{sec:conclusion}

In this paper, we compared CLM and MLM for text generation across medical, movie synopsis, and authorship verification datasets. We explored different masking approaches to determine the most effective ones. We then assessed the quality of the generations; first through quantitative and qualitative evaluations, and then by assessing their usability on three downstream tasks. Our results established that MLM models could generate better texts than CLM models.

\subsection{Limitations}
\label{sec:limitations}
Our findings provide valuable insights into the performance of MLM and CLM approaches for text generation. However, in this section, we want to acknowledge some of the limitations that may have impacted the generalisability of our results. \\


\textbf{Quantitative Evaluation Metrics:}
The quantitative evaluation relied upon n-gram-based metrics because they provide a valuable indication of text quality. However, it should be noted that they were initially designed for tasks such as machine translation and text summarization. As such, these measures may not fully align with our kind of texts. In addition, it could be that MLM models had an advantage in this evaluation as part of the true tokens were already provided to the models.

\textbf{Length of the Texts:}
The BERT family of models, including those we employed, have a maximum input sequence length of 512 tokens. This may limit the application of MLM-based text generation in domains that usually involve longer texts.

\textbf{Size of the Models:}
Due to computational resources and the fact that encoder-only models are generally smaller, we had to restrict our study to models that have hundreds of millions of parameters. Therefore, their performance may be limited compared to larger state-of-the-art models with billions of parameters. Further research should focus on that instead.

\subsection{Future Work}
Our comparison of MLM and CLM for text generation opened up multiple directions for future research. In this section, we outline four potential developments that could expand on our paper. \\

\textbf{Creative Writing:}
Future work could involve our method to generate more unstructured texts in tasks such as creative writing. This would demonstrate whether our approach could also enhance generations' creativity, on top of their quality.

\textbf{Iterative Refinement:}
Iterative refinement \cite{lee-etal-2018-deterministic} involves gradually improving the initial generations over multiple iterations until good text quality is achieved. When applied to MLM, this process could entail identifying and masking problematic tokens or text sequences, and then re-generating those parts to improve the quality of the final output.

\textbf{More Downstream Tasks:}
This paper showed that texts of higher quality do not necessarily lead to higher performance in the studied downstream tasks. As we focused on NER, text classification and authorship verification; future work should include a broader range of NLP tasks, such as sentiment analysis, question answering and POS tagging. By evaluating the performance of models trained on MLM-generated data across a diverse set of tasks, we will be able to show the versatility and robustness of this method with more certainty.

\textbf{On Medical Data:}
MLM models showed the best performance compared to CLM in the medical dataset. Therefore, future research could further investigate the application of MLM for data augmentation and synthetic data generation in the biomedical domain. This domain particularly benefits from such approaches due to the challenges of accessing data because of privacy concerns \cite{belkadi2023generating}. Future research could involve additional types of texts and masking strategies, as well as evaluating the generated data on a range of medical downstream tasks. Another critical aspect would be to collaborate with clinicians to assess the plausibility of the generated texts along with their quality. This could provide us valuable insights into the possible use of this approach in clinical settings.

    \section*{Acknowledgements}
We thank Prof Juntao Li from Soochow University's NLP group for the valuable feedback on our work.

    \section*{author contributions}
NM: experiments and writing (the original thesis, which this article is based upon); SB: writing; LH: article structure and writing (extracting from the thesis); GN: supervision.
\clearpage

\bibliography{nodalida2023}

\bibliographystyle{IEEEtran}

\end{document}